\documentclass{article}
\usepackage{ijcai22}

\usepackage{times}
\usepackage{soul}
\usepackage{url}
\usepackage[hidelinks]{hyperref}
\usepackage[utf8]{inputenc}
\usepackage[small]{caption}
\usepackage{graphicx}
\usepackage{amsmath}
\usepackage{amsthm}
\usepackage{algorithm}
\usepackage{algorithmic}
\usepackage{multirow}
\usepackage{tabularx}
\newtheorem{problem}{Problem}
\newtheorem{definition}{Definition}[section]
\usepackage{balance}
\newcolumntype{P}[1]{>{\centering\arraybackslash}p{#1}}
\usepackage{arydshln}
\usepackage{booktabs,siunitx}
\usepackage{bm}
\usepackage{pifont}
\usepackage{balance}
\usepackage{amsfonts}
\usepackage{nicematrix}

\usepackage[T1]{fontenc}
\usepackage[utf8]{inputenc}
\usepackage{babel}
\usepackage[font=small,labelfont=bf]{caption}
\usepackage{hyperref}
\usepackage{xcolor,soul,lipsum}

\usepackage{graphicx}
\usepackage{capt-of}% or \usepackage{caption}
\usepackage{booktabs}
\usepackage{varwidth}
\newsavebox\tmpbox
\DeclareCaptionLabelFormat{andfigure}{#1~#2  \&  \figurename~\thefigure}

\makeatletter
\def\adl@drawiv#1#2#3{%
        \hskip.5\tabcolsep
        \xleaders#3{#2.5\@tempdimb #1{1}#2.5\@tempdimb}%
                #2\z@ plus1fil minus1fil\relax
        \hskip.5\tabcolsep}
\newcommand{\cdashlineCustom}[1]{%
  \noalign{\vskip\aboverulesep
           \global\let\@dashdrawstore\adl@draw
           \global\let\adl@draw\adl@drawiv}
  \cdashline{#1}
  \noalign{\global\let\adl@draw\@dashdrawstore
           \vskip\belowrulesep}}
\makeatother

\title{Recipe2Vec: Multi-modal Recipe Representation Learning \\ with Graph Neural Networks}

\author{
Yijun Tian$^1$,
Chuxu Zhang$^{2*}$,
Zhichun Guo$^1$,
Yihong Ma$^1$,
Ronald Metoyer$^1$,
Nitesh V. Chawla$^{1*}$
\affiliations
$^1$Department of Computer Science, University of Notre Dame, USA\\ 
$^2$Department of Computer Science, Brandeis University, USA\\
\emails
$^1$\{yijun.tian, zguo5, yma5, rmetoyer, nchawla\}@nd.edu,
$^2$chuxuzhang@brandeis.edu
}
\begin{document}
\maketitle
\def\thefootnote{*}\footnotetext{Corresponding authors}\def\thefootnote{\arabic{footnote}}

% $$$$$$$$$$$$$$$$$$$$$$$$$$$$$$$$$$$$$$$$$$$$$$$$$$$$$$$$$$
\begin{abstract}
Learning effective recipe representations is essential in food studies. Unlike what has been developed for image-based recipe retrieval or learning structural text embeddings, the combined effect of multi-modal information (i.e., recipe images, text, and relation data) receives less attention. In this paper, we formalize the problem of \textit{multi-modal recipe representation learning} to integrate the visual, textual, and relational information into recipe embeddings. In particular, we first present \textit{Large-RG}, a new recipe graph data with over half a million nodes, making it the largest recipe graph to date. We then propose \textit{Recipe2Vec}, a novel graph neural network based recipe embedding model to capture multi-modal information. Additionally, we introduce an adversarial attack strategy to ensure stable learning and improve performance. Finally, we design a joint objective function of node classification and adversarial learning to optimize the model. Extensive experiments demonstrate that \textit{Recipe2Vec} outperforms state-of-the-art baselines on two classic food study tasks, i.e., cuisine category classification and region prediction. Dataset and codes are available at \url{https://github.com/meettyj/Recipe2Vec}.
\end{abstract}

% -------------- Pipeline ----------------
\begin{figure}[ht]
	\centering
	\includegraphics[width=0.9\columnwidth]{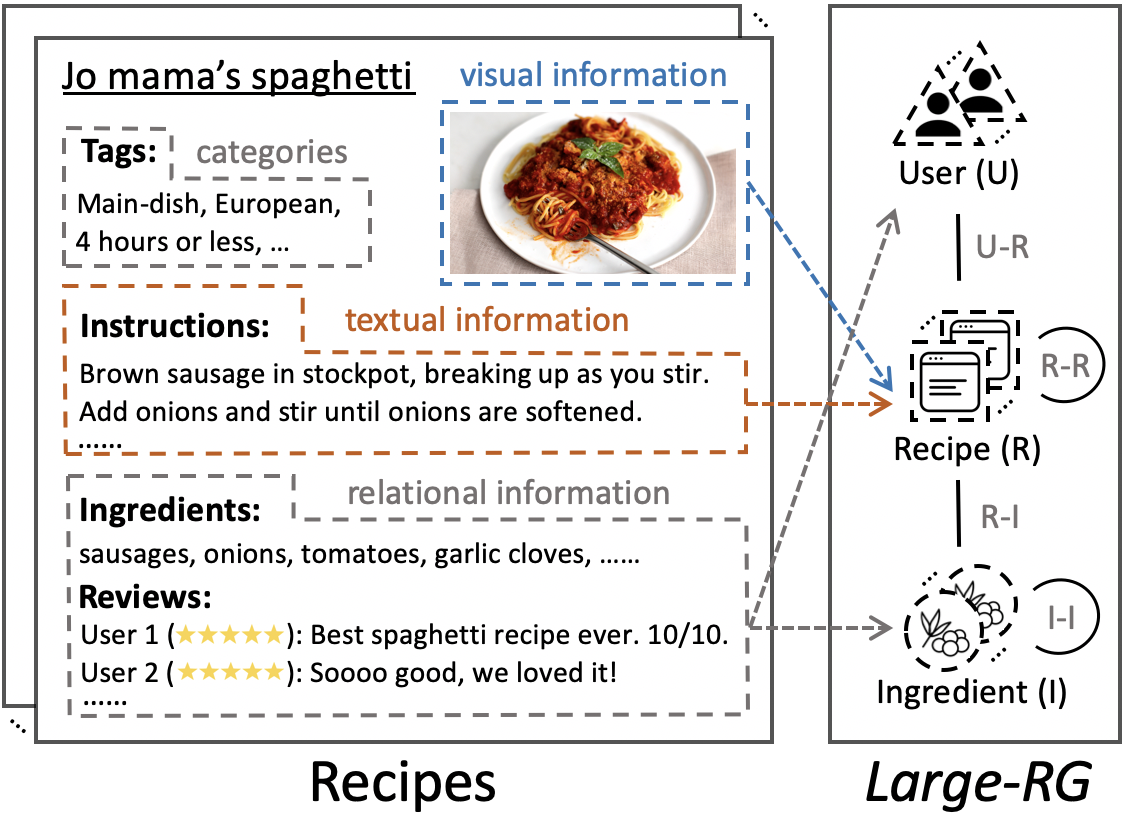}
	\caption{The illustration of \textit{Large-RG}. 
	We associate visual and textual information with recipe nodes, and extract various types of relations among user, recipe, and ingredient nodes.
	}
	\label{fig:intro}
\end{figure}

% $$$$$$$$$$$$$$$$$$$$$$$$$$$$$$$$$$$$$$$$$$$$$$$$$$$$$$$$$$
\section{Introduction}

% brief intro
Large-scale food data provides a wealth of information about food and can assist in resolving many critical societal issues \cite{min2019survey}. Recipe representation learning, in particular, embeds recipes in a latent space, allowing for the extraction of hidden information from massive food datasets and facilitating various application tasks that rely on a robust food space representation, such as culinary habits analysis \cite{culinary_habits_2}, recipe healthiness estimation \cite{Rokicki2018TheIO}, recipe recommendation \cite{frontiers_recipe_rec}, and recipe classification \cite{reciptor}. 

% why our work
Existing approaches to recipe representation learning rely entirely on textual content \cite{reciptor}. They take advantage of the recipe ingredients and instructions but overlook the relationship between recipes and food items. Besides leveraging the textual information, rn2vec \cite{rn2vec} proposes a recipe network embedding model to capture the relational information. However, rn2vec ignores the large number of images associated with each recipe and thus results in suboptimal performance. While some studies consider recipe images, they typically focus on image-based applications such as recognizing and retrieving recipes from images \cite{DBLP:journals/corr/abs-1709-09429,retrieval_attention_1}. These works attempt to align the images and the text in a shared embedding space for specific downstream tasks, but ignore the unique information contained in each modality.

% problem and dataset
In this paper, we propose the problem of \textit{multi-modal recipe representation learning}, which leverages different modalities such as images, text, and relations to learn recipe embeddings. To solve this problem, we first create and release a new recipe graph, \textit{Large-RG}, which contains over half a million nodes and is the largest recipe graph to date. As shown in Fig. \ref{fig:intro}, we extract visual, textual, and relational information of each recipe to build \textit{Large-RG}, which consists of three types of nodes and four types of edges. We then propose \textit{Recipe2Vec}, a novel graph neural network (GNN) based model for multi-modal recipe representation learning. Specifically, we first encode the node attributes using various pre-trained neural networks, including a two-stage LSTM \cite{recipe1m_1} for instructions and a ResNet \cite{resnet} for images. Next, we introduce a multi-view neighbor sampler to capture both local and high-order information from graph schema-based and meta-path-based neighbors, respectively. We also design several neural network based modules to fuse the information from different nodes, modalities, and relations. In addition, we introduce a feature-based adversarial attack strategy to ensure stable learning and improve model performance. Finally, we design a novel combined objective function of node classification and adversarial learning to optimize the model. To summarize, our major contributions are as follows:
\begin{itemize}
  \item As the first attempt to study the problem of \textit{multi-modal recipe representation learning}, we create and release \textit{Large-RG}, a new recipe graph data with over half a million nodes, which is the largest recipe graph to date.
  
  \item We propose a novel GNN-based recipe embedding model \textit{Recipe2Vec} to solve the problem. \textit{Recipe2Vec} is able to capture visual, textual, and relational information and learn effective representations through several neural network modules. We further introduce a novel objective function to optimize the model.

  \item We conduct extensive experiments to evaluate the performance of our model. The results show the superiority of \textit{Recipe2Vec} by comparing it with state-of-the-art baselines on two classic food study tasks: cuisine category classification and region prediction.
\end{itemize}

% $$$$$$$$$$$$$$$$$$$$$$$$$$$$$$$$$$$$$$$$$$$$$$$$$$$$$$$$$$
\section{Related Work}
This work is closely related to recipe datasets, recipe representation learning, and graph neural networks.

\noindent
\textbf{Recipe Datasets.} 
Existing datasets focus exclusively on either recipe images, text, or the simple relationship between ingredients \cite{recipe1m_1,foodkg,rn2vec,flavorGraph}, but fail to contain all of these information and further neglect the complex structure and user information, resulting in poor generality and integrity. Different from existing works, we create and release a recipe graph data which contains extensive relational information and multi-modal information including images and text.

\noindent
\textbf{Recipe Representation Learning.}
Existing works focus on using textual content to learn recipe representations \cite{reciptor,rn2vec}. They take advantage of the instructions and ingredients associated with recipes but ignore the large number of recipe images. Several other works consider the images, \cite{DBLP:journals/corr/abs-1804-11146,recipe1m_1,DBLP:journals/corr/abs-1905-01273,recipe_retrieval_1} but they typically concentrate on the recipe-image retrieval task and attempt to align images and text together in a shared space, resulting in information loss for both modalities. Instead, we focus on learning recipe representations using multi-modal information extracted from images, text, and relations.

\noindent
\textbf{Graph Neural Networks.}
Many GNNs \cite{gat,rgcn,han,hetgnn-kdd19,zhang2019shne} were proposed to learn vectorized node embeddings. They take advantage of both node attributes and relational information to learn node embeddings for different graph mining tasks. For example, GAT \cite{gat} employs self-attention mechanisms to measure the impacts of different neighbors and combine their effects to obtain node embeddings. HAN \cite{han} uses hierarchical attentions to learn embeddings by aggregating information from both node-level and semantic-level structures. Inspired by these studies, we build a GNN-based model to learn recipe embeddings. In addition, adversarial learning has shown great performance in training neural networks \cite{adversarial_1,adversarial_2}. We thus introduce a feature-based adversarial attack strategy to ensure stable learning and further improve performance.

% -------------- Data Statistic ----------------
\begin{table}[t]
\begin{minipage}[]{0.69\columnwidth}
\centering
\hypertarget{fig:pie}{} 
\captionlistentry[figure]{}
\captionsetup{labelformat=andfigure}
\captionof{table}{The statistics of \textit{Large-RG}, cuisine categories, and region categories.
}
\resizebox{\linewidth}{!}{
\begin{NiceTabular}[b]{c|c|c}
    \toprule
    Component & Name & Number \\
    \midrule
    \multirow{3}{*}{Nodes} 
     & user & 38,624 \\
     & recipe & 472,515\\
     & ingredient & 22,186 \\
    \midrule
    \multirow{4}{*}{Edges} 
     & user-recipe & 1,193,179 \\
     & recipe-recipe & 644,256 \\
     & recipe-ingredient & 4,440,820 \\
     & ingredient-ingredient & 170,642 \\
    \bottomrule
\end{NiceTabular}
}
\label{tab:data_statistic}
\end{minipage}
\hfill
\begin{minipage}[]{0.3\linewidth}
\centering
\includegraphics[width=25mm]{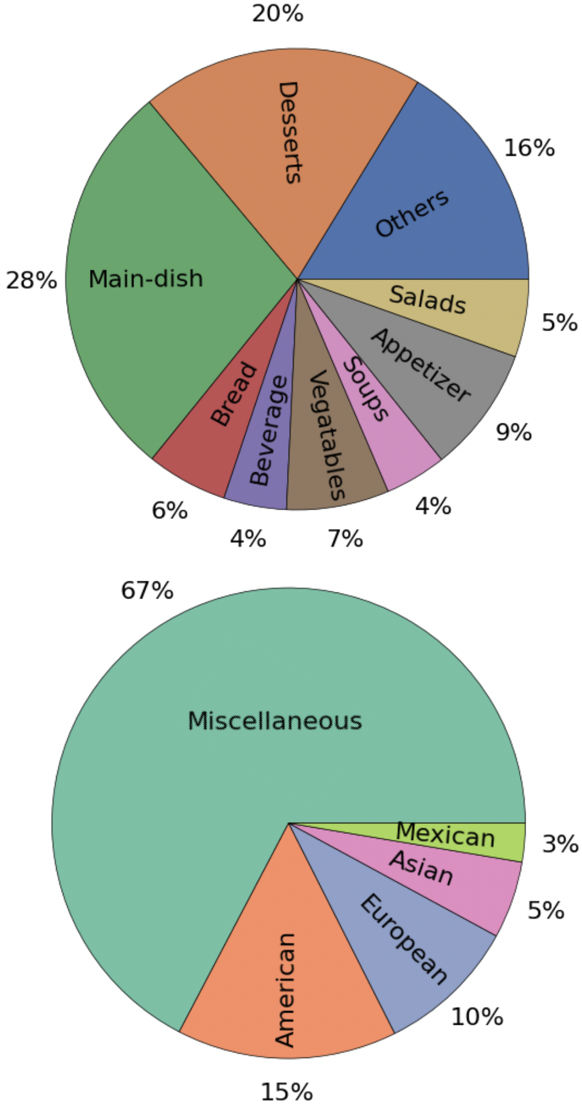}
\end{minipage}
\end{table}

% $$$$$$$$$$$$$$$$$$$$$$$$$$$$$$$$$$$$$$$$$$$$$$$$$$$$$$$$$$
\section{Preliminary}

In this section, we describe the concept of heterogeneous recipe graph and formally define the problem of \textit{multi-modal recipe representation learning}. We further introduce the \textit{Large-RG} data that we create in this work. 

\begin{definition}
{\bf Heterogeneous Recipe Graph.} 
A heterogeneous recipe graph is defined as a graph $G = (V, E, X)$ with multiple types of nodes $V$ (i.e., User, Recipe, Ingredient) and edges $E$ (e.g., U-R, R-R, R-I, I-I in Fig. \ref{fig:intro}). In addition, nodes are associated with attributes $X$, e.g., images, instructions, and nutrients.
\end{definition}

\begin{problem}
{\bf Multi-modal Recipe Representation Learning.}
Given a recipe graph $G = (V, E, X)$, the task is to design a learning model $\mathcal{F}_\Theta$ with parameters $\Theta$ to learn $d$-dimensional recipe embeddings $\mathcal{E} \in \mathbb{R}^d$, while encoding the multi-modal information (i.e., visual, textual, and relational information). The learned recipe representations can be utilized in various downstream tasks such as cuisine category classification and region prediction.
\end{problem}

% -------------- Pipeline ----------------
\begin{figure*}[ht]
	\centering
	\includegraphics[width=\textwidth]{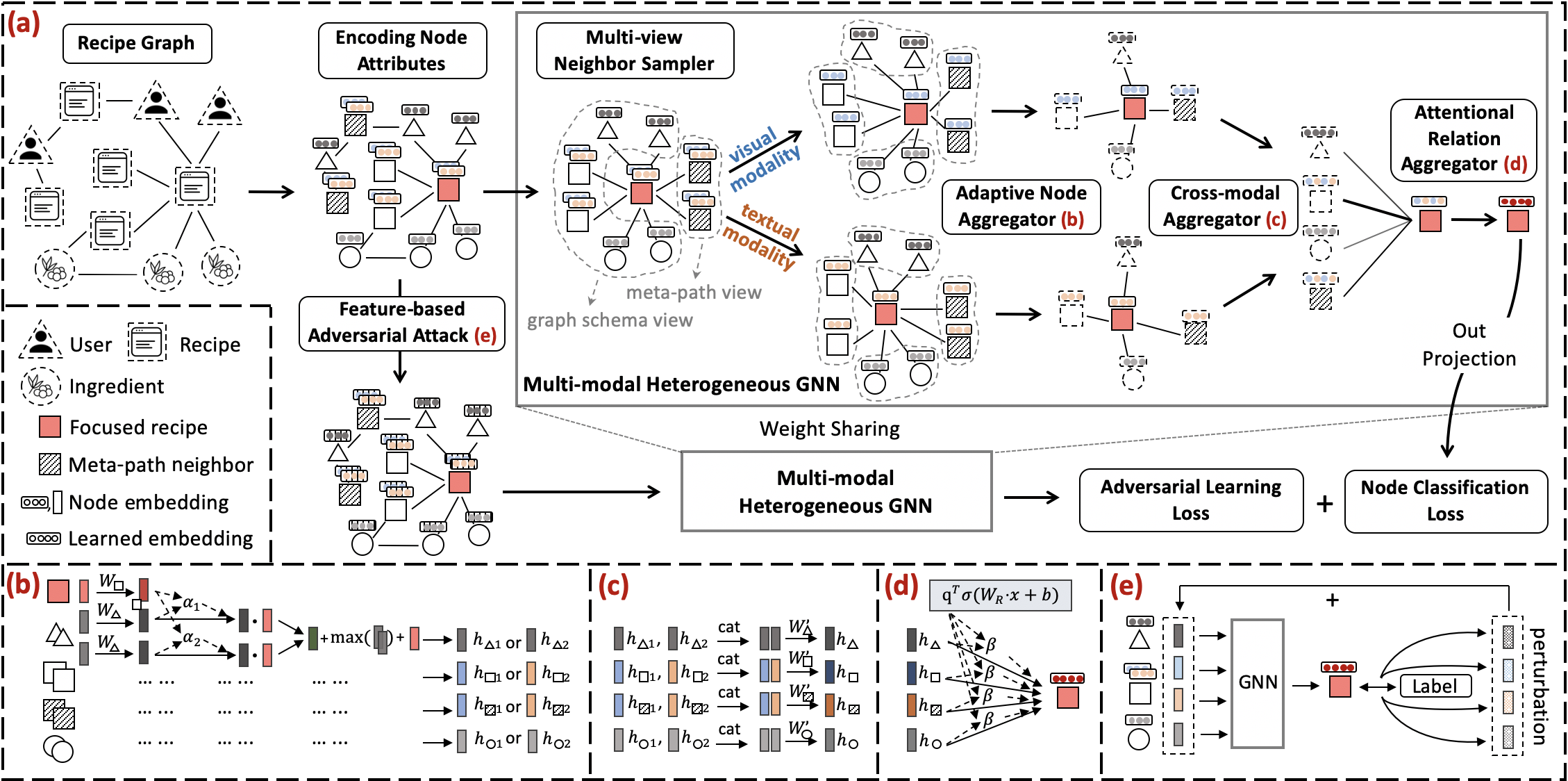}
	\caption{(a) The overall framework of \textit{Recipe2Vec}: we first encode the node attributes and propose a multi-modal heterogeneous GNN with various modules to encode visual, textual, and relational information for learning recipe embeddings. 
	We then introduce a feature-based adversarial attack strategy to ensure stable learning and design a joint loss of node classification and adversarial learning to optimize the model;
	(b) Adaptive node aggregator: encoding information from nodes under a specific relation; 
	(c) Cross-modal aggregator: fusing information from different modalities;
	(d) Attentional relation aggregator: using attention mechanism to fuse embeddings over different relations;
	(e) Feature-based adversarial attack: learning adversarial features by generating adversarial perturbations for input features.
	}
	\label{fig:framework}
\end{figure*}

\noindent
\textbf{Large-RG Data.}
\label{sec_data}
To solve the problem, we first create \textit{Large-RG} data. The statistics of \textit{Large-RG} are reported in Tab. \ref{tab:data_statistic} and Fig. \hyperlink{fig:pie}{2}. Specifically, we first collect recipes from Recipe1M \cite{recipe1m_1} and crawl the user ratings for each recipe from \textit{food.com}. We then match each ingredient to the USDA nutritional database \cite{usda_sr} to determine its nutritional value. After that, we build \textit{Large-RG} by representing users, recipes, and ingredients as nodes and connecting them with four types of relations. In particular, we build the recipe-ingredient relation by connecting each recipe and its ingredients. We construct the user-recipe relation based on the user ratings. We also connect recipe nodes by the similarity determined by FoodKG \cite{foodkg} and connect ingredient nodes based on the co-occurring probabilities \cite{flavorGraph}. Additionally, we crawl recipe tags from food.com to create two new categories and assign each recipe a cuisine class and a region class.

% $$$$$$$$$$$$$$$$$$$$$$$$$$$$$$$$$$$$$$$$$$$$$$$$$$$$$$$$$$
\section{Model}

In this section, we present the details of \textit{Recipe2Vec}.
As illustrated in Fig. \ref{fig:framework}, we first encode the node attributes and develop a multi-modal heterogeneous GNN to learn recipe embeddings. The proposed GNN contains several modules including a multi-view neighbor sampler, an adaptive node aggregator (Fig. \ref{fig:framework}(b)), a cross-modal aggregator (Fig. \ref{fig:framework}(c)), and an attentional relation aggregator (Fig. \ref{fig:framework}(d)). We also introduce a feature-based adversarial attack strategy (Fig. \ref{fig:framework}(e)) to ensure stable learning. Finally, we design a novel objective function to optimize the model.

% ------------------------------
\subsection{Encoding Nodes Attributes}
We pretrain a ResNet \cite{resnet} to encode the recipe images and a two-stage LSTM \cite{recipe1m_1} to encode the instructions. Next, we associate the pre-trained image embeddings ($x_{img}$) and instruction embeddings ($x_{ins}$) with recipe nodes as the node attributes. Similarly, we consider nutrients as the attributes of ingredient nodes ($x_{ing}$) and random vectors as the attributes of user nodes ($x_{user}$). Due to the various modality of attributes, we introduce a modality-specific input projection to project the node attributes into a shared embedding space:
\begin{equation}
h_{i,m} = W_{m} \cdot x_{i,m},
\label{h_im}
\end{equation}
where $W_{m} \in \mathbb{R}^{{d_{m}} \times d}$ is the projection for modality $m$, $x_{i,m}$ is the associated attribute of node $v_i$ with dimension $d_{m}$, and $h_{i,m}$ is the projected feature of node $v_i$ with dimension $d$.

% ------------------------------
\subsection{Multi-modal Heterogeneous GNN}
For each recipe node $v_i$ with input feature $h_{i,m}$, we first introduce a multi-view neighbor sampler to sample nodes from graph schema view and meta-path view. We then use an adaptive node aggregator to encode the information from nodes under a specific relation. Next, we propose a cross-modal aggregator to fuse the information from different modalities. After that, we design an attentional relation aggregator to combine the information from different relations and obtain the final recipe embedding. 

% ------------------------------
\noindent
\textbf{Multi-view Neighbor Sampler.} 
We introduce the multi-view neighbor sampler to 
simultaneously capture local information through graph schema view and high-order information through meta-path view \cite{metapath2vec}. In the graph schema view, the node aggregates the information from its direct neighbors, which we denote as the graph schema-based neighbors $N_i^S$. In the meta-path view, the neighbors are determined by meta-path walks for the high-order information. In particular, given a meta-path $\mathcal{P}$ that starts from $v_i$, instead of taking all the nodes that appear in the meta-path as neighbors, we define an importance pooling to select the most important $p$ nodes. We start by simulating multiple meta-path walks from $v_i$ and compute the $L_1$ normalized visit count for each node. We then determine the important neighbors by choosing the top $p$ nodes with the highest normalized visit counts. We denoted these selected important neighbors as the meta-path-based neighbors $N_i^P$. Finally, the neighbor nodes of $v_i$ is the combined set of $N_i^S$ and $N_i^P$, denoted as $N_i$. In addition, since the recipe nodes are associated with different modalities (e.g., visual and textual), we split the graph into modality-specific graphs to account for each modality individually.

% ------------------------------
\noindent
\textbf{Adaptive Node Aggregator.} 
Given a node $v_i$ and its neighbor nodes $N_i$, we first sample the relation-specific neighbors $N_{i,r} \subset N_i$ by choosing the nodes that connect to $v_i$ through relation $r$. Then, we calculate the unnormalized attention score $e_{ij,m}$ between $v_i$ and $v_j \in N_{i,r}$ given the modality $m$, and normalize the $e_{ij,m}$ using $softmax$ function to make the coefficients easily comparable: 
\begin{equation}
\begin{aligned}
e_{ij,m} &= \text{LeakyReLU}\left[W_{ij} \cdot (h_{i,m}\|h_{j,m})\right],\\
\alpha_{ij,m} &= \frac{\exp(e_{ij,m})}{\sum_{k\in N_{i,r}}^{}\exp(e_{ik,m})},
\end{aligned}
\end{equation}
where $\|$ indicates concatenation, $W_{ij}\in \mathbb{R}^{2d \times d}$ is a shared weight vector, and $\alpha_{ij,m}$ is the attention weight between $v_i$ and $v_j$ for modality $m$. We then use the $\alpha_{ij,m}$ as coefficients to linearly combine the interaction and yield the united interaction-dependent affinity $A_m$ for each modality $m$:
\begin{equation}
A_m = \sum_{j\in N_{i,r}} \alpha_{ij,m} \cdot W_a \cdot (h_{i,m} \odot h_{j,m}),
\end{equation}
where $W_a \in \mathbb{R}^{d \times d}$ is a shared trainable weight and $\odot$ denotes the element-wise product. We then aggregate the neighbor node features following GIN \cite{gin} while including two shared transformation matrix $W_{i}\in \mathbb{R}^{d \times d}$ and $W_{j}\in \mathbb{R}^{d \times d}$ to enhance learning capacity. We further make it adaptive to specific downstream tasks by incorporating the interaction-dependent affinity $A_m$, which makes the message dependent on the affinity between $v_i$ and $v_j$:
\begin{equation}
h_{i,r,m} = W_o \cdot [W_i \cdot h_i + \mathrm{max}(\sum_{j\in N_{i,r}} W_j \cdot h_j) + A_m],
\end{equation}
where $h_{i,r,m}$ is the encoded embedding of $v_i$ through relation $r$ for modality $m$, and $W_o \in \mathbb{R}^{d \times d}$ is a shared weight vector.

% ------------------------------
\noindent
\textbf{Cross-modal Aggregator.} 
To better exploit the information in different modalities and learn a joint embedding, we design the cross-modal aggregator to fuse embeddings from visual and textual modalities. Specifically, we first concatenate the $h_{i,r,image}$ and $h_{i,r,text}$, and then introduce a node type-specific projection $W_{\phi_i} \in \mathbb{R}^{2d \times d}$ to transform the concatenated embedding, with $\phi_i$ are the node type of $v_i$:
\begin{equation}
h_{i,r} = W_{\phi_i} \cdot (h_{i,r, image}\|h_{i,r,text}),
\end{equation}
where $h_{i,r}$ is the learned cross-modal embedding of node $v_i$ through relation $r$.

% ---------------- cuision table ---------------- 
\begin{table*}[ht]
\caption{Cuisine category classification results.}
\begin{center}
\resizebox{0.8\linewidth}{!}{
\begin{NiceTabular}{cc|P{1cm}P{1cm}P{1cm}P{1cm}P{1cm}P{1cm}P{1.0cm}P{1.0cm}P{1.0cm}P{1cm}}
    \toprule
    Metric & Method & \multicolumn{1}{c}{Appetizer} & \multicolumn{1}{c}{Beverage} & \multicolumn{1}{c}{Bread} & \multicolumn{1}{c}{Soups} & \multicolumn{1}{c}{Salads} & \multicolumn{1}{c}{Desserts} & \multicolumn{1}{c}{Vegetables} & \multicolumn{1}{c}{Main-dish} & \multicolumn{1}{c}{Others} & \multicolumn{1}{c}{Total}\\
    \midrule
    
    %  --- F1 --- 
    \multirow{11}{*}{F1} 
    & \multirow{1}{*}{TextCNN} 
    & 50.4 & 85.9 & 73.8 & 62.3 & 68.9 & 86.9 & 49.1 & 75.9 & 42.4 & 73.5 \\
    & \multirow{1}{*}{ResNet} 
    & 41.9 & 88.5 & 74.3 & 75.9 & 76.8 & 88.7 & 64.3 & 83.4 & 55.6 & 78.7 \\
    & \multirow{1}{*}{mp2vec} 
    & 50.0 & 83.2 & 67.3 & 57.7 & 63.2 & 85.9 & 54.9 & 75.7 & 43.0 & 72.4 \\
    & \multirow{1}{*}{RGCN} 
    & 52.6 & 81.8 & 65.3 & 52.5 & 60.1 & 84.0 & 49.4 & 73.7 & 43.5 & 70.7 \\
    & \multirow{1}{*}{GAT} 
    & 53.2 & 84.7 & 60.7 & 51.2 & 64.9 & 84.4 & 38.5 & 73.5 & 41.8 & 71.0 \\
    & \multirow{1}{*}{GIN} 
    & 68.4 & \underline{91.9} & 82.4 & 81.2 & 81.4 & 91.4 & 70.9 & 86.1 & 70.3 & 83.9 \\
    & \multirow{1}{*}{HAN} 
    & 67.0 & 90.4 & 78.3 & 79.1 & 78.9 & 89.4 & 66.1 & 84.6 & 67.3 & 81.8 \\
    & \multirow{1}{*}{HGT} 
    & 49.1 & 89.3 & 75.1 & 79.1 & 78.0 & 88.4 & 63.5 & 82.8 & 55.3 & 79.2 \\
    & \multirow{1}{*}{Reciptor} 
    & 68.2 & 89.1 & 79.3 & 77.4 & 71.4 & 90.2 & 59.9 & 84.1 & 57.4 & 80.9 \\
    & \multirow{1}{*}{rn2vec} 
    & \underline{76.8} & 91.4 & \underline{84.4} & \underline{83.7} & \underline{83.9} & \underline{91.5} & \underline{76.8} & \underline{86.9} & \underline{73.5} & \underline{85.9} \\
    \cdashlineCustom{2-12}
    & \multirow{1}{*}{\textit{Recipe2Vec}} 
    & \textbf{82.3} & \textbf{95.5} & \textbf{89.0} & \textbf{87.9} & \textbf{87.2} & \textbf{94.5} & \textbf{82.3} & \textbf{89.4} & \textbf{78.2} & \textbf{89.4} \\
    \midrule
    
    %  --- Accuracy --- 
    \multirow{11}{*}{Acc} 
    & \multirow{1}{*}{TextCNN} 
    & 45.5 & 86.3 & 74.2 & 61.0 & 70.9 & 90.8 & 41.8 & 85.8 & 35.2 & 69.3 \\
    & \multirow{1}{*}{ResNet} 
    & 32.0 & 89.8 & 72.9 & 67.9 & 72.9 & 92.1 & 59.9 & 90.3 & 56.6 & 74.9 \\
    & \multirow{1}{*}{mp2vec} 
    & 40.7 & 87.8 & 61.3 & 68.4 & 61.8 & 90.1 & 50.4 & 86.4 & 34.7 & 68.6 \\
    & \multirow{1}{*}{RGCN} 
    & 47.3 & 78.7 & 57.8 & 43.8 & 58.0 & 92.1 & 44.6 & 83.7 & 37.1 & 66.9 \\
    & \multirow{1}{*}{GAT} 
    & 44.4 & 81.4 & 47.6 & 39.6 & 70.3 & 94.6 & 26.0 & 84.1 & 40.0 & 66.4 \\
    & \multirow{1}{*}{GIN} 
    & 66.0 & \underline{93.8} & 79.7 & 78.3 & 81.4 & 92.0 & 65.6 & 89.7 & 69.5 & 81.7 \\
    & \multirow{1}{*}{HAN} 
    & 62.0 & 91.7 & 75.2 & 75.1 & 80.4 & 92.2 & 61.6 & 89.1 & 64.3 & 79.5 \\
    & \multirow{1}{*}{HGT} 
    & 40.7 & 89.4 & 70.4 & 75.7 & 75.9 & 93.4 & 54.6 & 90.8 & 53.2 & 75.5 \\
    & \multirow{1}{*}{Reciptor} 
    & 62.7 & 86.2 & 76.2 & 75.4 & 71.3 & 93.6 & 55.0 & 91.2 & 52.7 & 77.4 \\
    & \multirow{1}{*}{rn2vec} 
    & \underline{72.5} & 90.4 & \underline{81.3} & \underline{80.3} & \underline{82.7} & \underline{94.3} & \underline{72.9} & \underline{91.4} & \underline{70.5} & \underline{84.0} \\
    \cdashlineCustom{2-12}
    & \multirow{1}{*}{\textit{Recipe2Vec}} 
    & \textbf{80.1} & \textbf{95.8} & \textbf{87.5} & \textbf{87.5} & \textbf{88.8} & \textbf{95.5} & \textbf{80.1} & \textbf{91.6} & \textbf{76.0} & \textbf{87.6} \\
    \bottomrule

\end{NiceTabular}}
\end{center}
\label{tab:cuisine_results}
\end{table*}

% ---------------- region table ---------------- 
\begin{table}[ht]
\caption{Region prediction results.}
\begin{center}
\resizebox{\columnwidth}{!}{
\begin{NiceTabular}{cc|P{1.05cm}P{1.05cm}P{1.05cm}P{1.05cm}P{1.05cm}}
    \toprule
    Metric & Method & \multicolumn{1}{c}{American} & \multicolumn{1}{c}{European} & \multicolumn{1}{c}{Asian} & \multicolumn{1}{c}{Mexican} & \multicolumn{1}{c}{Total}\\
    \midrule
    
    %  --- F1 --- 
    \multirow{11}{*}{F1} 
    & \multirow{1}{*}{TextCNN} 
    & 67.8 & 56.7 & 64.3 & 37.8 & 62.2 \\
    & \multirow{1}{*}{ResNet} 
    & 69.7 & 58.1 & 66.7 & 40.1 & 64.1 \\
    & \multirow{1}{*}{mp2vec} 
    & 70.2 & 64.0 & \underline{81.4} & 20.1 & 67.7 \\
    & \multirow{1}{*}{RGCN} 
    & 70.3 & 64.1 & 70.7 & 30.3 & 66.4 \\
    & \multirow{1}{*}{GAT} 
    & 69.4 & 63.6 & 80.3 & 24.3 & 67.0 \\
    & \multirow{1}{*}{GIN} 
    & 73.7 & 65.5 & 75.7 & 55.4 & 70.2 \\
    & \multirow{1}{*}{HAN} 
    & 71.8 & 66.8 & 78.7 & 58.3 & 70.2 \\
    & \multirow{1}{*}{HGT} 
    & 71.8 & 65.5 & 78.1 & 50.5 & 69.5 \\
    & \multirow{1}{*}{Reciptor} 
    & 73.2 & 66.8 & 76.2 & 57.2 & 70.6 \\
    & \multirow{1}{*}{rn2vec} 
    & \underline{75.6} & \underline{68.5} & 79.6 & \underline{59.8} & \underline{73.0} \\
    \cdashlineCustom{2-7}
    & \multirow{1}{*}{\textit{Recipe2Vec}} 
    & \textbf{78.7} & \textbf{73.8} & \textbf{85.3} & \textbf{68.9} & \textbf{77.5} \\
    \midrule
    
    %  --- Accuracy --- 
    \multirow{11}{*}{Acc} 
    & \multirow{1}{*}{TextCNN} 
    & 72.3 & 57.0 & 60.0 & 28.0 & 62.2 \\
    & \multirow{1}{*}{ResNet} 
    & 75.8 & 55.8 & 63.2 & 30.1 & 64.1 \\
    & \multirow{1}{*}{mp2vec} 
    & 75.4 & 65.9 & 77.5 & 11.5 & 67.7 \\
    & \multirow{1}{*}{RGCN} 
    & 76.7 & 68.6 & 57.1 & 18.8 & 66.4 \\
    & \multirow{1}{*}{GAT} 
    & 73.5 & 68.2 & 73.0 & 14.3 & 67.0 \\
    & \multirow{1}{*}{GIN} 
    & 77.1 & 66.2 & 69.3 & 48.2 & 70.2 \\
    & \multirow{1}{*}{HAN} 
    & 71.7 & 73.0 & 70.7 & 50.7 & 70.2 \\
    & \multirow{1}{*}{HGT} 
    & 75.2 & 68.5 & 70.5 & 38.9 & 69.5 \\
    & \multirow{1}{*}{Reciptor} 
    & 72.6 & \underline{69.4} & \underline{77.8} & 49.7 & 70.6 \\
    & \multirow{1}{*}{rn2vec} 
    & \underline{78.1} & 67.4 & \underline{77.8} & \underline{55.0} & \underline{73.0} \\
    \cdashlineCustom{2-7}
    & \multirow{1}{*}{\textit{Recipe2Vec}} 
    & \textbf{78.6} & \textbf{74.0} & \textbf{83.8} & \textbf{71.0} & \textbf{77.5} \\
    \bottomrule

\end{NiceTabular}
}
\end{center}
\label{tab:region_results}
\end{table}

% ------------------------------
\noindent
\textbf{Attentional Relation Aggregator.} 
We further introduce the attentional relation aggregator to learn the importance of each relation and fuse them accordingly. Specifically, we first apply a weight matrix $W_R \in \mathbb{R}^{d \times d}$ to transform the $h_{i,r}$ and use a relation-level vector $q \in \mathbb{R}^{d}$ to calculate the similarity. We then average the similarity scores of all nodes that connected to relation $r$ to obtain the importance score $w_{i,r}$ for node $v_i$. The process is formulated as follows:
\begin{equation}
w_{i,r} =\frac{1}{|V_r|}\sum_{i \in V_r} q^\text{T} \cdot \tanh(W_{R} \cdot h_{i,r}+b),
\end{equation}
where $V_r$ denotes the set of nodes that are connected to relation $r$, and $b \in \mathbb{R}^{d}$ is the bias vector. Next, we normalize $w_{i,r}$ to get the final relation-level attention weight $\beta_{i,r}$:
\begin{equation}
\beta_{i,r}=\frac{\exp(w_{i,r})}{\sum_{r \in R_i} \exp(w_{i,r})},
\end{equation}
where $R_i$ indicates the associated relations of $v_i$. After that, we fuse the cross-modal embedding $h_{i,r}$ with $\beta_{i,r}$ to obtain the final recipe embedding $h_i$:
\begin{equation}
h_{i}=\sum_{r=1}^{R_i} \beta_{i,r} \cdot h_{i,r}.
\label{h_i}
\end{equation}

% ------------------------------
\subsection{Objective Function}
The learned recipe embedding $h_i$ can be used for various downstream tasks such as cuisine category classification, region prediction, or recipe recommendation. Specifically, in this work, we can introduce a supervised node classification loss (e.g., cross-entropy) to classify the cuisine categories (or predict the recipe regions):
\begin{equation}
L_{sup}=-\sum_{i \in \mathcal{Y}_{T}} Y_i \log (FC(h_i)),
\end{equation}
where $\mathcal{Y}_T$ is the set of training data,
$FC$ denotes the fully connected layer,
and $Y_i$ is the one-hot label of $v_i$.

% ------------------------------
\subsection{Feature-based Adversarial Learning}
Since the above objective function only considers the explicit feature information associated with each node, we further introduce an adversarial attack strategy to leverage the regularization power of adversarial features. Specifically, we choose Projected Gradient Descent \cite{pgd} as the default attacker to generate adversarial features on the fly. Compared to the vanilla training that we send original input features $h_{i,m}$ (Eq.~\ref{h_im}) into GNN to obtain the learned recipe embedding $h_i$ (Eq.~\ref{h_i}), adversarial training takes the maliciously perturbed features $h_{i,m}^\prime$ as input and obtains the perturbed recipe embedding $h_i^\prime$, which is further used to calculate the adversarial learning loss $L_{adv}$:
\begin{equation}
\begin{aligned}
h_{i,m}^\prime &= h_{i,m} + \epsilon_m; 
h_{i}^\prime = GNN (h_{i,m}^\prime),\\
L_{adv} &= \max_{\epsilon_m \in \mathbb{S}} [ -\sum_{i \in \mathcal{Y}_{T}} Y_i \log (FC(h_i^\prime))],
\end{aligned}
\end{equation}
where $\epsilon_m$ is the adversarial perturbation for modality $m$ and $\mathbb{S}$ is the allowed perturbation range. The final objective function $L$ is defined as the weighted combination of $L_{sup}$ and $L_{adv}$: 
\begin{equation}
L = L_{sup} + \lambda L_{adv},
\end{equation}
where $\lambda$ is a trade-off weight for balancing two losses.

% ---------------- Error cases ---------------- 
\begin{table*}[t]
\setcounter{table}{4}
\caption{Error cases for cuisine category classification.}
\begin{center}
\resizebox{0.85\linewidth}{!}{
\begin{NiceTabular}{c|c|c|cccc}
    \toprule
    Recipe Title & Ingredients & Ground Truth & TextCNN & HAN & rn2vec & \textit{Recipe2Vec} \\
    \midrule
    
    \multirow{1}{*}{Corn and Coriander Soup} & corn, capsicum, onion, coriander, ... & \multirow{1}{*}{Soups} & \multirow{1}{*}{Main-dish} & \multirow{1}{*}{Soups} & \multirow{1}{*}{Soups} & \multirow{1}{*}{Soups} \\
    \midrule
    
    \multirow{1}{*}{Herbed Tomatoes} & tomatoes, herb, onion, rosemary, ... & \multirow{1}{*}{Vegetables} & \multirow{1}{*}{Main-dish} & \multirow{1}{*}{Main-dish} & \multirow{1}{*}{Vegetables} & \multirow{1}{*}{Vegetables} \\
    \midrule
    
    \multirow{1}{*}{Zucchini Spoon Bread} & zucchini, eggs, milk, almonds, ... & \multirow{1}{*}{Bread} & \multirow{1}{*}{Vegetables} & \multirow{1}{*}{Main-dish} & \multirow{1}{*}{Vegetables} & \multirow{1}{*}{Bread} \\
    \midrule
    
    \multirow{1}{*}{Spaghetti With Pea Sauce} & spaghetti, butter, shallots, peas, ... & \multirow{1}{*}{Main-dish} & \multirow{1}{*}{Appetizer} & \multirow{1}{*}{Salads} & \multirow{1}{*}{Salads} & \multirow{1}{*}{Main-dish} \\
    \bottomrule
\end{NiceTabular}
}
\end{center}
\label{tab:case_study}
\end{table*}

% -------------- ablation study ----------------
\begin{table}[t]
\setcounter{table}{3}
\caption{F1 scores of different model variants.}
\begin{center}
\resizebox{\columnwidth}{!}{
\begin{NiceTabular}{cc|cccccc}
    \toprule
    Task & Category & -- NS & -- NA & -- CA & -- RA & -- AL & \textit{Recipe2Vec}\\
    \midrule
    
    %  --- Cuisine --- 
    \multirow{10}{*}{\shortstack{Cuisine\\category\\classification}} & \multirow{1}{*}{Appetizer} 
    & 81.9 & 79.0 & 78.5 & 81.5 & 81.2 & \textbf{82.3} \\
    & \multirow{1}{*}{Beverage} 
    & 95.4 & 94.4 & 94.0 & 95.0 & 95.1 & \textbf{95.5} \\
    & \multirow{1}{*}{Bread} 
    & 88.6 & 87.4 & 86.9 & 88.6 & 88.3 & \textbf{89.0} \\
    & \multirow{1}{*}{Soups} 
    & 87.5 & 86.1 & 86.0 & 87.8 & 87.7 & \textbf{87.9} \\
    & \multirow{1}{*}{Salads} 
    & 86.5 & 85.6 & 85.7 & 86.5 & 86.5 & \textbf{87.2} \\
    & \multirow{1}{*}{Desserts} 
    & 94.1 & 93.2 & 92.9 & 94.2 & 94.1 & \textbf{94.5} \\
    & \multirow{1}{*}{Vegetables} 
    & 81.3 & 78.2 & 78.6 & 81.9 & 81.1 & \textbf{82.3} \\
    & \multirow{1}{*}{Main-dish} 
    & 89.1 & 87.6 & 87.7 & 89.3 & 89.0 & \textbf{89.4} \\
    & \multirow{1}{*}{Others} 
    & 77.3 & 75.2 & 75.9 & 78.1 & 78.0 & \textbf{78.2} \\
    & \multirow{1}{*}{Total}
    & 88.9 & 87.4 & 87.3 & 89.1 & 88.8 & \textbf{89.4} \\
    \midrule
    
    %  --- Region --- 
    \multirow{5}{*}{\shortstack{Region\\prediction}} & \multirow{1}{*}{American} 
    & 78.5 & 77.3 & 78.2 & 78.4 & 78.5 & \textbf{78.7} \\
    & \multirow{1}{*}{European} 
    & 73.0 & 72.0 & 66.7 & 73.7 & 73.3 & \textbf{73.8} \\
    & \multirow{1}{*}{Asian} 
    & 84.3 & 83.1 & 82.9 & 84.4 & 83.7 & \textbf{85.3} \\
    & \multirow{1}{*}{Mexican} 
    & 65.6 & 64.4 & 64.5 & 67.2 & 67.1 & \textbf{68.9} \\
    & \multirow{1}{*}{Total} 
    & 76.8 & 75.7 & 75.1 & 77.1 & 76.9 & \textbf{77.5} \\
    \bottomrule
    
\end{NiceTabular}
}
\end{center}
\label{tab:ablation_studies}
\end{table}

% $$$$$$$$$$$$$$$$$$$$$$$$$$$$$$$$$$$$$$$$$$$$$$$$$$$$$$$$$$
\section{Experiments}
In this section, we conduct extensive experiments to evaluate the performance of \textit{Recipe2Vec} and show related analyses.

% ------------------------------
\subsection{Baseline Methods}
We compare with 10 baselines including the classic classifiers \textbf{TextCNN} \cite{textcnn}, \textbf{ResNet} \cite{resnet}, homogeneous graph embedding models \textbf{GAT} \cite{gat}, \textbf{GIN} \cite{gin}, heterogeneous graph embedding models \textbf{mp2vec} \cite{metapath2vec}, \textbf{RGCN} \cite{rgcn}, \textbf{HAN} \cite{han}, \textbf{HGT} \cite{hgt}, and recipe representation learning models \textbf{Reciptor} \cite{reciptor} and \textbf{rn2vec} \cite{rn2vec}.

% ------------------------------
\subsection{Implementation Details}
We split the data into train/validation/test set by 70/15/15. For the proposed \textit{Recipe2Vec}, we set the learning rate to 0.005, the hidden size to 128, the input dimension of instruction and image embeddings to 512, the input dimension of ingredient embeddings to 46, batch size to 4096, meta-path $\mathcal{P}$ to recipe-user-recipe, the number of meta-path neighbors $p$ to 10, training epochs to 100, the trade-off factor $\lambda$ to 0.1, the perturbation range $\mathbb{S}$ to 0.02, number of iterations for attack to 5, and the attack step size to 0.005.

% ------------------------------
\subsection{Performance Comparison}
We use Micro-F1 and Accuracy (Acc) as the evaluation metrics and report the performances on cuisine category classification and region prediction tasks in Tab. \ref{tab:cuisine_results} and Tab. \ref{tab:region_results}, respectively. The best and second-best values are highlighted by bold and underline. According to these tables, we can find that \textit{Recipe2Vec} outperforms all the baselines for both tasks in all cases. Specifically, classic classifiers (i.e., TextCNN and ResNet) perform poorly because of the neglect of both relational information and multi-modal information. Graph embedding methods (e.g., GIN and HAN) obtain decent performance after incorporating the complex relational information. Similarly, recipe representation learning models (e.g., rn2vec) achieve satisfactory results, but they fail to encode the influence of different modalities. Finally, \textit{Recipe2Vec} achieves the best performance compared to all the baselines, by improving \textbf{+3.5\%} (F1) and \textbf{+3.6\%} (Acc) in cuisine category classification, and \textbf{+4.5\%} (F1 and Acc) in region prediction. This demonstrates that \textit{Recipe2Vec} can obtain better recipe embeddings compared to other models.

% ------------------------------
\subsection{Ablation Study}
Since \textit{Recipe2Vec} contains various essential components (i.e., multi-view neighbor sampler (NS), adaptive node aggregator (NA), cross-modal aggregator (CA), attentional relation aggregator (RA), and feature-based adversarial learning (AL)), we conduct ablation studies to analyze the contributions of different components by removing each of them independently (see Tab. \ref{tab:ablation_studies}). Specifically, removing NA and CA significantly affects the performance, showing that both NA and CA have large contributions to \textit{Recipe2Vec}. In addition, we remove NS, RA, and AL from our model, respectively. The decreasing performance of these model variants demonstrates the effectiveness of NS, RA, and AL in enhancing the model. Finally, \textit{Recipe2Vec} achieves the best results in all cases, indicating the strong capability of different components in our model.

% ------------------------------
\subsection{Case Study}
To show the performance of different models with concrete examples, we analyze the misclassified cases in the cuisine category classification task, as shown in Tab. \ref{tab:case_study}. Specifically, TextCNN misclassifies all of these four recipes, indicating that the textual features cannot fully represent the recipes and that ignoring relational information may lead to suboptimal performance. HAN successfully classifies the recipe \textit{Corn and Coriander Soup}, but fails to classify the others. One potential reason is that relying solely on meta-path-based neighbors may result in information loss. rn2vec successfully classifies the recipes \textit{Corn and Coriander Soup} and \textit{Herbed Tomatoes}, but fails to classify the other two. This may be because the model cannot fully capture the multi-modal information by only using a simple GNN-based structure. However, our model \textit{Recipe2Vec} takes into account the visual, textual, and relational information through several neural network modules, which clearly distinguishes the difference among categories and correctly classifies these recipes.

% ------------------------------
\subsection{Embedding Visualization}
\begin{figure}
	\centering
	\includegraphics[width=0.85\columnwidth]{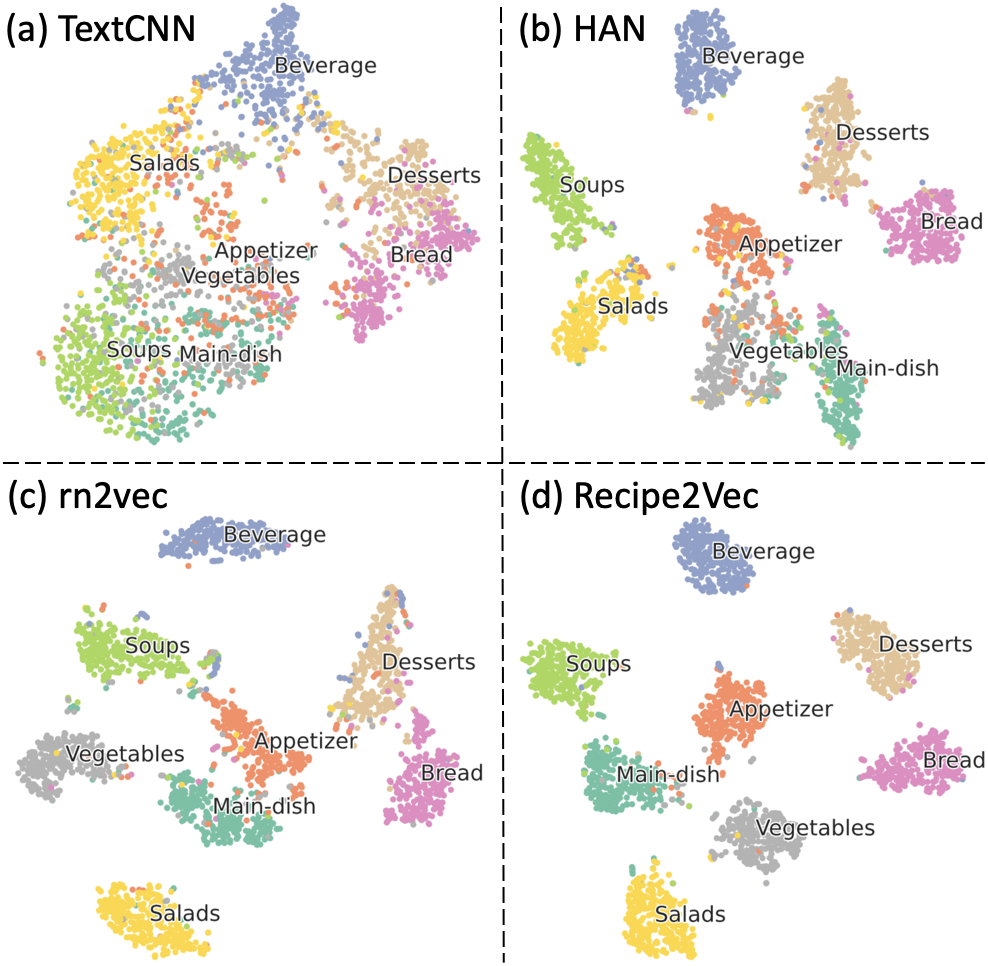}
	\caption{
	Embedding visualization of recipes.
	}
	\label{fig:visualization}
\end{figure}

For a more intuitive understanding and comparison, we visualize embeddings of different models using t-SNE. As shown in Fig. \ref{fig:visualization}, TextCNN does not perform well. Only Beverage, Salads, Desserts, and Bread are separated apart while other categories are mixed. HAN can separate most categories but fails to distinguish the Appetizer, Vegetables, and Main-dish. While rn2vec can successfully separate all categories, their distinctions are obscure, i.e., points from different categories are close to each other. However, our model \textit{Recipe2Vec} can clearly identify each category and distinctly separate all of them. This again demonstrates that \textit{Recipe2Vec} can learn discriminative recipe embeddings.

% $$$$$$$$$$$$$$$$$$$$$$$$$$$$$$$$$$$$$$$$$$$$$$$$$$$$$$$$$$
\section{Conclusion}
In this paper, we propose and formalize the problem of \textit{multi-modal recipe recommendation learning}. To solve this problem, we create and release \textit{Large-RG}, a new and large-scale recipe graph data to facilitate graph-based food studies. Furthermore, we develop \textit{Recipe2Vec}, a novel GNN-based recipe embedding model. \textit{Recipe2Vec} is able to capture visual, textual, and relational information through various carefully designed neural network modules. We also design a joint objective function of node classification and adversarial learning to optimize the model. Extensive experiments show that \textit{Recipe2Vec} outperforms state-of-the-art baselines on two classic food study tasks.

% --- Acknowledgements ---
\section*{Acknowledgements}
This work is supported by the Agriculture and Food Research Initiative grant no. 2021-67022-33447/project accession no.1024822 from the USDA National Institute of Food and Agriculture.

%% The file named.bst is a bibliography style file for BibTeX 0.99c
\bibliographystyle{named}
\bibliography{main}

\end{document}